\documentclass{svproc}
\usepackage{url}

\usepackage{needspace} 

\usepackage{graphicx} 
\usepackage{caption} 
\usepackage{subcaption} 
\usepackage{multicol} 
\usepackage[bottom]{footmisc} 
\usepackage{amsmath} 
\usepackage{amssymb}  
\usepackage{siunitx} 

\usepackage{import}
\usepackage{tabularx}                   
\usepackage[table]{xcolor} 
\usepackage[numbers]{natbib}     

\usepackage{tikz}
\usepackage{pgf}
\usepackage{pgffor}
\usepgfmodule{shapes}
\usepgfmodule{plot}
\usetikzlibrary{decorations}
\usetikzlibrary{snakes}
\usetikzlibrary{patterns}
\usetikzlibrary{backgrounds}
\usepackage{circuitikz}

\DeclareMathOperator{\PWM}{PWM}

\DeclareMathOperator{\pwm}{\textsc{pwm}}
\DeclareMathOperator{\bemf}{\textsc{bemf}}
\DeclareMathOperator{\sign}{sign}
\DeclareMathOperator{\diag}{diag}
\newcommand{\f}[1]{\textsc{#1}}

\DeclareSIUnit\degpersec{\deg\per\second}
\DeclareSIUnit\pwmmu{pwm.mach.units}
\DeclareSIUnit\pwm{\textsc{pwm}}

\newtheorem{assumption}{Assumption}
\numberwithin{assumption}{chapter}

\newtheorem{axiom}{Axiom}
\numberwithin{axiom}{chapter}

\usetikzlibrary{arrows.meta}

\makeatletter

\pgfdeclarearrow{
  name = ipe _linear,
  defaults = {
    length = +1bp,
    width  = +.666bp,
    line width = +0pt 1,
  },
  setup code = {
    \pgfarrowssetbackend{0pt}
    \pgfarrowssetvisualbackend{
      \pgfarrowlength\advance\pgf@x by-.5\pgfarrowlinewidth}
    \pgfarrowssetlineend{\pgfarrowlength}
    \ifpgfarrowreversed
      \pgfarrowssetlineend{\pgfarrowlength\advance\pgf@x by-.5\pgfarrowlinewidth}
    \fi
    \pgfarrowssettipend{\pgfarrowlength}
    \pgfarrowshullpoint{\pgfarrowlength}{0pt}
    \pgfarrowsupperhullpoint{0pt}{.5\pgfarrowwidth}
    \pgfarrowssavethe\pgfarrowlinewidth
    \pgfarrowssavethe\pgfarrowlength
    \pgfarrowssavethe\pgfarrowwidth
  },
  drawing code = {
    \pgfsetdash{}{+0pt}
    \ifdim\pgfarrowlinewidth=\pgflinewidth\else\pgfsetlinewidth{+\pgfarrowlinewidth}\fi
    \pgfpathmoveto{\pgfqpoint{0pt}{.5\pgfarrowwidth}}
    \pgfpathlineto{\pgfqpoint{\pgfarrowlength}{0pt}}
    \pgfpathlineto{\pgfqpoint{0pt}{-.5\pgfarrowwidth}}
    \pgfusepathqstroke
  },
  parameters = {
    \the\pgfarrowlinewidth,%
    \the\pgfarrowlength,%
    \the\pgfarrowwidth,%
  },
}

\pgfdeclarearrow{
  name = ipe _pointed,
  defaults = {
    length = +1bp,
    width  = +.666bp,
    inset  = +.2bp,
    line width = +0pt 1,
  },
  setup code = {
    \pgfarrowssetbackend{0pt}
    \pgfarrowssetvisualbackend{\pgfarrowinset}
    \pgfarrowssetlineend{\pgfarrowinset}
    \ifpgfarrowreversed
      \pgfarrowssetlineend{\pgfarrowlength}
    \fi
    \pgfarrowssettipend{\pgfarrowlength}
    \pgfarrowshullpoint{\pgfarrowlength}{0pt}
    \pgfarrowsupperhullpoint{0pt}{.5\pgfarrowwidth}
    \pgfarrowshullpoint{\pgfarrowinset}{0pt}
    \pgfarrowssavethe\pgfarrowinset
    \pgfarrowssavethe\pgfarrowlinewidth
    \pgfarrowssavethe\pgfarrowlength
    \pgfarrowssavethe\pgfarrowwidth
  },
  drawing code = {
    \pgfsetdash{}{+0pt}
    \ifdim\pgfarrowlinewidth=\pgflinewidth\else\pgfsetlinewidth{+\pgfarrowlinewidth}\fi
    \pgfpathmoveto{\pgfqpoint{\pgfarrowlength}{0pt}}
    \pgfpathlineto{\pgfqpoint{0pt}{.5\pgfarrowwidth}}
    \pgfpathlineto{\pgfqpoint{\pgfarrowinset}{0pt}}
    \pgfpathlineto{\pgfqpoint{0pt}{-.5\pgfarrowwidth}}
    \pgfpathclose
    \ifpgfarrowopen
      \pgfusepathqstroke
    \else
      \ifdim\pgfarrowlinewidth>0pt\pgfusepathqfillstroke\else\pgfusepathqfill\fi
    \fi
  },
  parameters = {
    \the\pgfarrowlinewidth,%
    \the\pgfarrowlength,%
    \the\pgfarrowwidth,%
    \the\pgfarrowinset,%
    \ifpgfarrowopen o\fi%
  },
}

\newdimen\ipeminipagewidth

\tikzstyle{ipe import} = [
  x=1bp, y=1bp,
  ipe node stretch/.store in=\ipenodestretch,
  ipe stretch normal/.style={ipe node stretch=1},
  ipe stretch normal,
  ipe node/.style={
    anchor=base west, inner sep=0, outer sep=0, scale=\ipenodestretch
  },
  ipe mark scale/.store in=\ipemarkscale,
  ipe mark tiny/.style={ipe mark scale=1.1},
  ipe mark small/.style={ipe mark scale=2},
  ipe mark normal/.style={ipe mark scale=3},
  ipe mark large/.style={ipe mark scale=5},
  ipe mark normal, 
  ipe circle/.pic={
    \draw[line width=0.2*\ipemarkscale]
      (0,0) circle[radius=0.5*\ipemarkscale];
    \coordinate () at (0,0);
  },
  ipe disk/.pic={
    \fill (0,0) circle[radius=0.6*\ipemarkscale];
    \coordinate () at (0,0);
  },
  ipe fdisk/.pic={
    \filldraw[line width=0.2*\ipemarkscale]
      (0,0) circle[radius=0.5*\ipemarkscale];
    \coordinate () at (0,0);
  },
  ipe box/.pic={
    \draw[line width=0.2*\ipemarkscale, line join=miter]
      (-.5*\ipemarkscale,-.5*\ipemarkscale) rectangle
      ( .5*\ipemarkscale, .5*\ipemarkscale);
    \coordinate () at (0,0);
  },
  ipe square/.pic={
    \fill
      (-.6*\ipemarkscale,-.6*\ipemarkscale) rectangle
      ( .6*\ipemarkscale, .6*\ipemarkscale);
    \coordinate () at (0,0);
  },
  ipe fsquare/.pic={
    \filldraw[line width=0.2*\ipemarkscale, line join=miter]
      (-.5*\ipemarkscale,-.5*\ipemarkscale) rectangle
      ( .5*\ipemarkscale, .5*\ipemarkscale);
    \coordinate () at (0,0);
  },
  ipe cross/.pic={
    \draw[line width=0.2*\ipemarkscale, line cap=butt]
      (-.5*\ipemarkscale,-.5*\ipemarkscale) --
      ( .5*\ipemarkscale, .5*\ipemarkscale)
      (-.5*\ipemarkscale, .5*\ipemarkscale) --
      ( .5*\ipemarkscale,-.5*\ipemarkscale);
    \coordinate () at (0,0);
  },
  /pgf/arrow keys/.cd,
  ipe arrow normal/.style={scale=1},
  ipe arrow tiny/.style={scale=.4},
  ipe arrow small/.style={scale=.7},
  ipe arrow large/.style={scale=1.4},
  ipe arrow normal,
  /tikz/.cd,
  ipe arrows/.style={
    ipe normal/.tip={
      ipe _pointed[length=1bp, width=.666bp, inset=0bp,
                   quick, ipe arrow normal]},
    ipe pointed/.tip={
      ipe _pointed[length=1bp, width=.666bp, inset=0.2bp,
                   quick, ipe arrow normal]},
    ipe linear/.tip={
      ipe _linear[length = 1bp, width=.666bp,
                  ipe arrow normal, quick]},
    ipe fnormal/.tip={ipe normal[fill=white]},
    ipe fpointed/.tip={ipe pointed[fill=white]},
    ipe double/.tip={ipe normal[] ipe normal},
    ipe fdouble/.tip={ipe fnormal[] ipe fnormal},
    ipe arc/.tip={ipe normal},
    ipe farc/.tip={ipe fnormal},
    ipe ptarc/.tip={ipe pointed},
    ipe fptarc/.tip={ipe fpointed},
  },
  ipe arrows, 
  ipe heavier/.style={line width=0.8},
  ipe fat/.style={line width=1.2},
  ipe ultrafat/.style={line width=2},
]

\tikzset{
  rgb color/.code args={#1=#2}{%
    \definecolor{tempcolor-#1}{rgb}{#2}%
    \tikzset{#1=tempcolor-#1}%
  },
}

\makeatother

\begin{document}
\mainmatter              
\title{Identification of Motor Parameters on Coupled Joints}
\titlerunning{Motor Torque and Joint Friction Identification}  
%
\author{Nuno Guedelha$^{1,2}$, Silvio Traversaro$^{1}$, Daniele Pucci$^{1}$}
\authorrunning{-} 
\institute{$^{1}$  iCub Facility, Istituto Italiano di Tecnologia, Genova, \\
Italy 16163 {\tt\small {firstname.lastname}@iit.it} \\
This project has received funding from the European Unions Horizon 2020 research and innovation programme under grant agreement No. 731540 (An.Dy). \\
$^{2}$ Universita degli Studi di Genova, DIBRIS
}

\maketitle              

\begin{abstract}
The estimation of the motor torque and friction parameters are crucial for implementing an efficient low level joint torque control. In a set of coupled joints, the actuators torques are mapped to the output joint torques through a coupling matrix, such that the motor torque and friction parameters appear entangled from the point of view of the joints. As a result, their identification is problematic when using the same methodology as for single joints. This paper proposes an identification method with an improved accuracy with respect to classical closed loop methods on coupled joints. The method stands out through the following key points: it is a direct open loop identification; it addresses separately each motor in the coupling; it accounts for the static friction in the actuation elements. The identified parameters should significantly improve the contribution of the feed-forward terms in the low level control of coupled joints with static friction.
\end{abstract}

\section{INTRODUCTION} \label{sec:introduction}

Some tasks require highly dynamic motions like running, jumping or even walking on uneven terrains, and for that, the robot has to account for the dynamic properties of the low level joint actuation. In typical torque control architectures composed by two nested control loops, this concern lies within the inner, low level control loop, which guaranties that the desired torque computed by the outer loop is generated at the joint level as expected within a delay that doesn't compromise the stability of the controller.

The feed-forward control allows to anticipate the changes in the controller setpoint, which in the case of low level joint torque control, would be the desired joint output torque. At that point, the only errors left to be corrected by the feedback controller are the model errors, the sensors noise and the external disturbancies applied to the system. This improves the stability of the controlled system, and the ability to use a wider range of gains in the feedback control depending on the desired stiffness of the joints. This type of control design is one among other model based control designs which benefit from the accurate identification of the joint low level actuation subsystem.

Commonly, identifying the feed-forward parameters goes through the breakdown of the overall transfer function relating the motor input parameter to the output torques, by modeling the sub-elements of the joint actuation system as well as their interaction. Throughout this paper, we will consider electrical joint actuation subsystems typically powered by a Direct Current (DC) motor, or a higher performance brush-less motor. It is actually the case for most of the humanoid robots $40 \si{\centi\metre}$ tall or above, apart from those equipped with hydraulic actuators.

\begin{figure}[!t]
\centering
\begin{subfigure}[t]{0.3\textwidth}
\centering
\includegraphics[height=4cm]{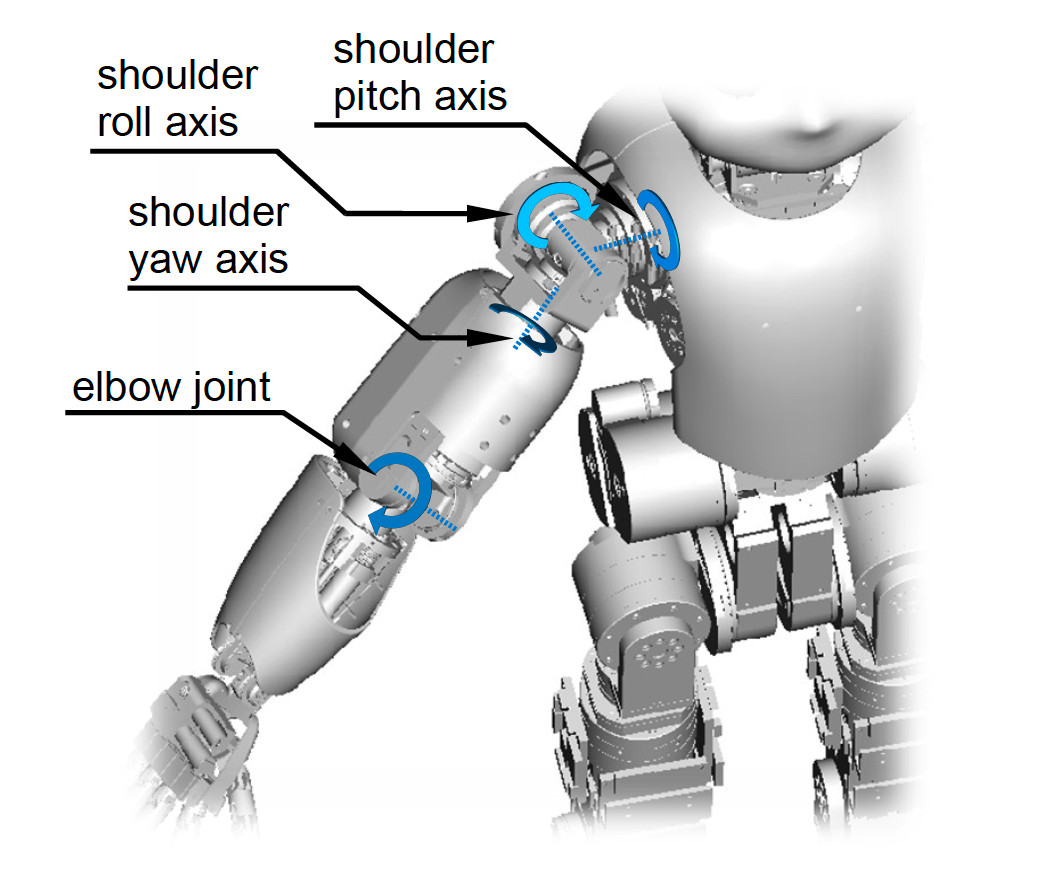}
\caption{\label{fig:shoulderJoints}}
\end{subfigure} \hspace{0.5cm}
\begin{subfigure}[t]{0.6\textwidth}
\centering
\includegraphics[height=4cm]{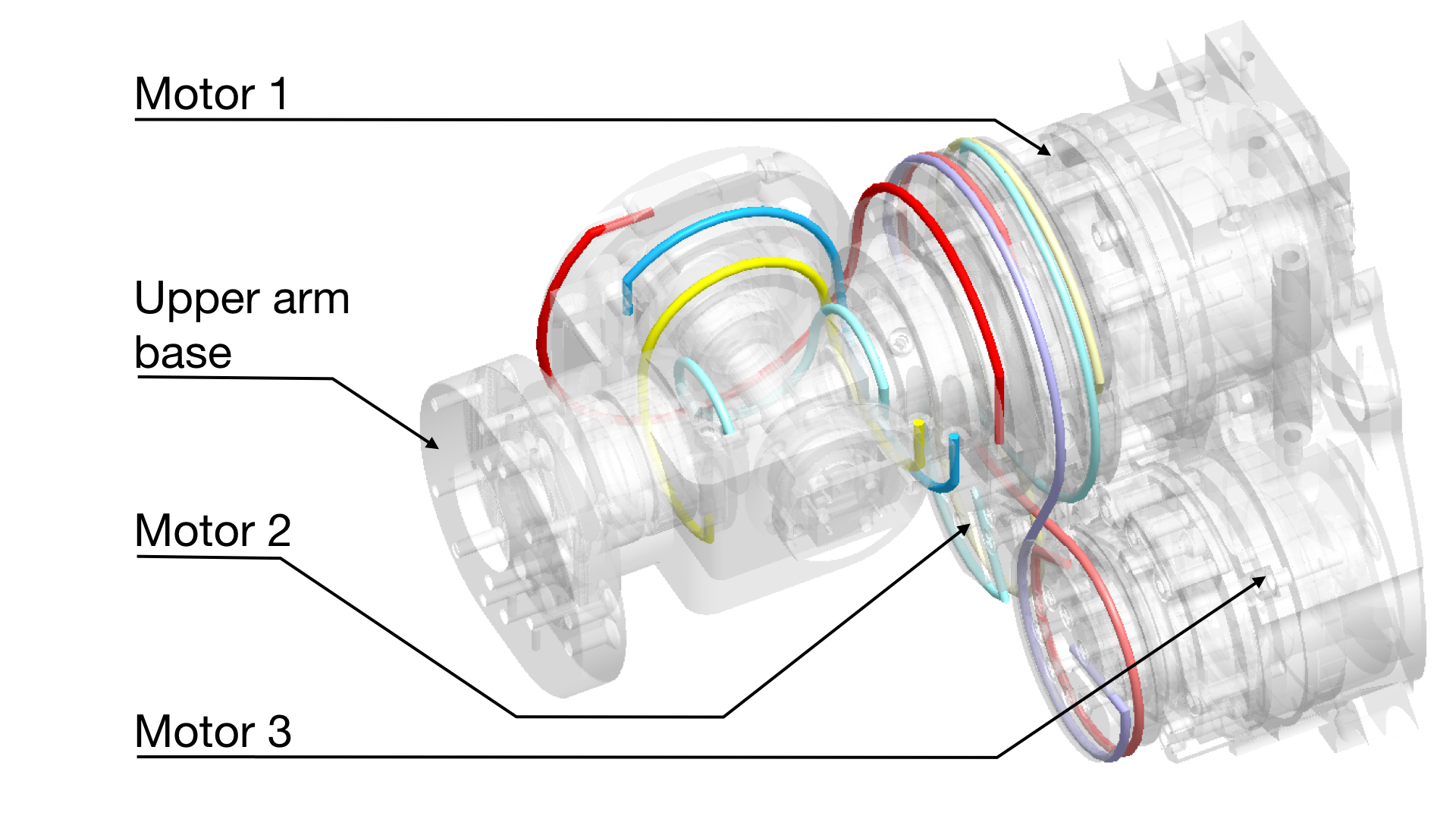}
\caption{}
\end{subfigure} \caption{CAD view of the iCub right arm with the three coupled shoulder joints (a) and CAD view of the shoudler joints actuated by the differential drive and three motors. Source: \cite[Fig.4 \& 5]{parmiggiani_design_2012}} \label{fig:shoulderCoupling}
\end{figure}

DC motors, or brush-less motors are typically driven by current or voltage PWM---Pulse Width Modulation---duty cycle. The generated torque is then multiplied in a reduction drive (gearbox or harmonic drive), then transmitted to the load on the output shaft, directly or through a coupling system, for instance as the one used on the humanoid robot iCub's torso or shoulders depicted in Figure \ref{fig:shoulderCoupling}. Some torque is wasted in the process, through friction in each of the transmission components. The output torque on the shaft is what we define as the joint torque. The identification process then consists in quantifying the parameters of the PWM to joint torque model.

A commonly used simple model defines the generated motor torque as a linear function of the motor input voltage duty cycle, and an overall friction term as an affine function of the joint velocity.

A large range of models have been employed to describe the friction dynamics, from static models like the Coulomb, the viscous and Stribeck models, to more complex ones, like the switching models (seven parameters model, Dahl model), or the most advanced dynamic models (LuGre model) accounting also for stick-slip phases transitions and other properties inherent to the interaction of surfaces in contact \citep{van2009study}.

More recent works proposed improvements on joint torque and external forces estimation, while the joints were not moving \citep{stolt_robotic_2015}, or were moving at very low velocities \citep{linderoth_robotic_2013}. The presented methods accounted for the significant torque disturbances due to static, Coulomb and viscous friction, modeling them either as a deterministic affine function \citep{stolt_sensorless_2015}, either as a uniformly distributed noise, with a range (the Coulomb "friction band") dependent on the velocity, combined to a Gaussian noise of zero mean, with a variance increasing with joint velocity \citep{linderoth_robotic_2013}. In both cases, the respective parameters were identified manually and few details were given on the procedure.
In \citep{popovic_modeling_1998}, a more thorough spectral-based identification is performed, accounting for a position dependent torque variations due to multiple parts in the transmission contributing to the overall friction.
This paper addresses the identification of the static and dynamic friction parameters on coupled joints, using a deterministic model, but avoids the complexity arising from the combination of the friction effects in the coupled motors and gearboxes. While addressing each motor transmission at a time, the paper focuses on the automatic identification the static friction, without any manual tuning as seen in previous works.
The paper is organized as follows: we first present an overview of the joint actuation architecture implemented on the major joints of iCub (the platform we used for performing our identification tests); after describing the model of each component composing one typical joint actuation sub-system, we build the model of the full transmission chain and pose the Newton-Euler equations describing its dynamics; based on those equations, we define the identification framework and methodology; we then address the case of the coupled joints and present the experiment results.

\section{BACKGROUND} \label{sec:background}

\subsection{Notation} \label{subsec:background_notation}

The notation used for describing the models and the algorithms in this paper is summarized as follows:

\begin{flushleft}
\begin{tabularx}{\columnwidth}{lX}
${}_\textsc{b}\tau_{m,l}$ & Motor torque applied on point $B$ of link $l$; \\
${}_\textsc{b}\tau_{f,l}$ & Generalized friction torque applied on point $B$ of link $l$; \\
${}_\textsc{b}\tau_{c,l}$ & Coulomb friction torque applied on point $B$ of link $l$; \\
${}_\textsc{b}\tau_{v,l}$ & Viscous friction torque applied on point $B$ of link $l$; \\
$F_n$ & Force normal to a contact surface; \\
$\mu,\sigma$ & Respectively Coulomb and viscous friction coefficients; \\
$\sign(\omega)$ & Sign of the angular velocity; \\
$\tau_J$ & Output joint torque (applied to the joint child link); \\
$\mathbb{I}_n$ & Identity matrix of dimension $n$.
\end{tabularx}
\end{flushleft}

\subsection{Friction Models}\label{subsec:friction_models}

While modeling the multiple components in the joint transmission chain, we will often come across friction torques that can be found in the bearings, reduction drives, coupling systems or result from electromotive forces.
Before addressing the friction modeling specific to each component, we present in this section the general concepts and typical models.
There is a fair list of candidate friction models: The most simple and common are the static ones (constant parameters and structure) and are not defined for a zero relative velocity between the two surfaces in contact; the "switching" models (seven parameters model, Dahl model described in \citep{van2009study}), and the most advanced dynamic models (LuGre model) switch between two distinct "stick" and "slip" states applying specific parameters for each of them; the dynamic models account for the transition between stick and slip phases in a continuous function.
We list and describe below the two most common.

\subsubsection*{The Static/Coulomb/Viscous friction model}

\begin{figure}[t]
\centering
\begin{subfigure}[t]{0.4\textwidth}
\centering
\includegraphics[width=\columnwidth]{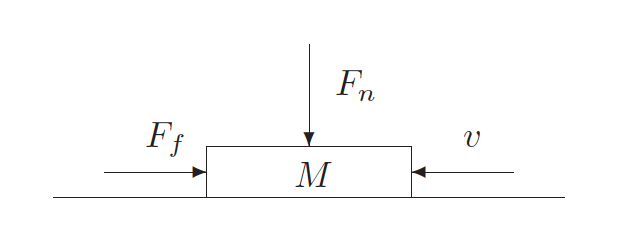}
\caption{} \label{fig:coulombViscousFrictionModelFlatSurface}
\end{subfigure} \hspace{0.5cm}
\begin{subfigure}[t]{0.4\textwidth}
\centering
\includegraphics[width=\columnwidth]{figs2/coulombViscousFrictionModelHinge.png}
\caption{} \label{fig:coulombViscousFrictionModelHinge}
\end{subfigure}
\caption{Representation of the friction force $F_f$ on an object M moving relative to a flat surface at velocity $v$ (a), or the friction torque $\tau_f$ on an object rotating about a rotary joint axis at velocity $\omega$ (b). $F_n$ is the force normal to rotation axis.} \label{fig:coulombViscousFrictionModel}
\end{figure}

We represent in Figure \ref{fig:coulombViscousFrictionModel} a moving object $M$, either sliding over a flat surface or revolving about a rotary joint (Figure \ref{fig:coulombViscousFrictionModelHinge}). In the rotary joint case, we can see the load force $F_n$ on the joint rotation axis ($F_n$ is normal to the axis), the rotation velocity $\omega$ and the total friction torque $\tau_f$ due to the rotary contact.

The friction torque can be written as a composition of three models: the Coulomb, the Viscous and the Static friction, as defined in \eqref{eq:staticCoulombViscousModel},
\begin{align}
\tau_f &= \tau_c + \tau_v + \tau_s \\
&= - \mu F_n \sign(\omega) - \sigma \omega - \mathcal{T}_s(\omega),
\end{align} \label{eq:staticCoulombViscousModel}
where the Coulomb friction $\tau_c$ is linear with respect to the load force $F_n$ on the rotation axis, and the viscous friction $\tau_v$ is a linear function of the axis angular velocity $\omega$. $\mu$ and $\sigma$ are respectively the Coulomb and viscous friction coefficients and are positive, such that the friction is always opposed to $\omega$ . The static friction $\tau_s$, initially expressed here as a general function of $\omega$, $\mathcal{T}_s(\omega)$, is the torque required to set the axis into motion starting from a null velocity, and typically is higher than the Coulomb and Viscous added components close to a null velocity. The first simple model would consider the static friction as a constant, usually higher than the Coulomb friction, as described in \cite{van2009study} and illustrated in Figure \ref{fig:staticStribeckFriction}(a). This approximation creates a discontinuity when the system velocity crosses zero, and thus can cause numerical issues and torque instabilities.

\subsubsection*{The Stribeck/Coulomb/Viscous friction model}

\begin{figure}[t]
\centering
\includegraphics[width=0.8\columnwidth]{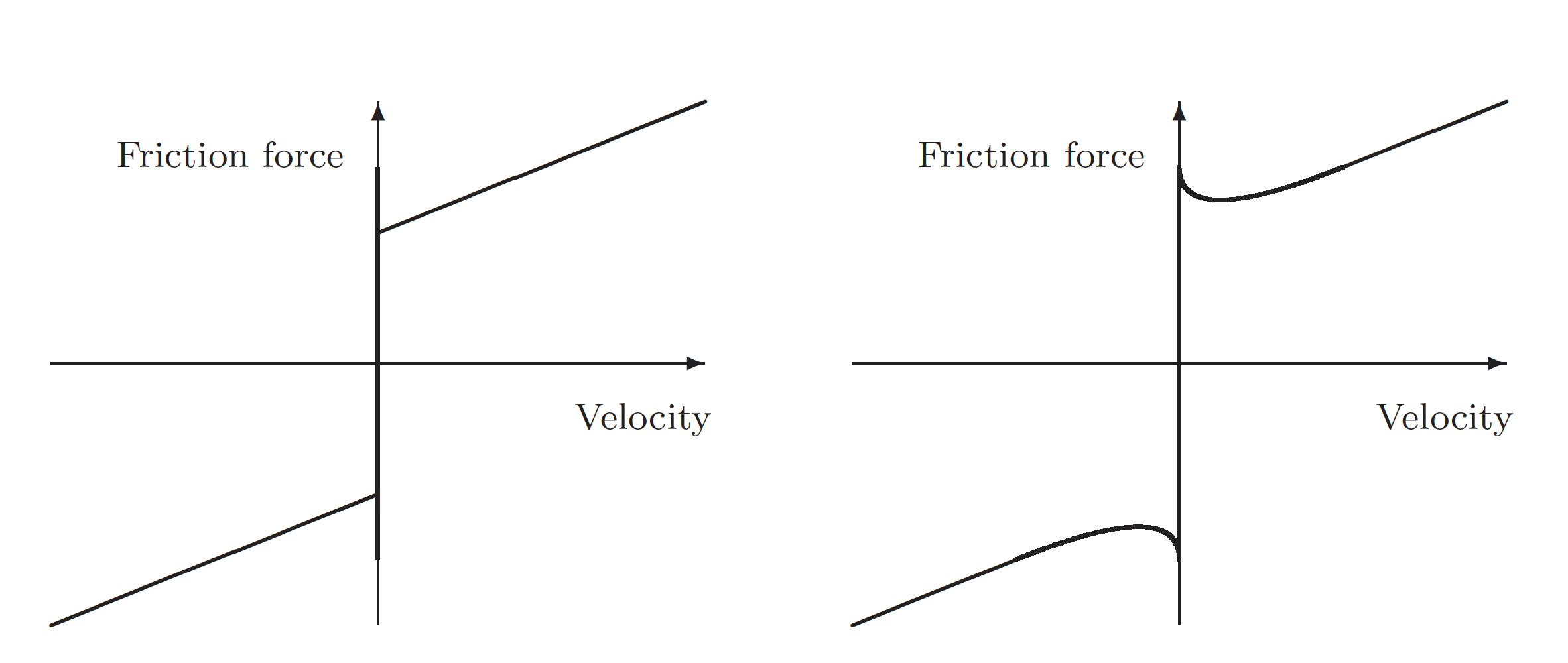}
\caption{Models of friction force versus angular velocity. (left) Static, Coulomb and viscous friction model. (right) Negative viscous, Coulomb and viscous friction model (Stribeck). Source: \cite{van2009study}.} \label{fig:staticStribeckFriction}
\end{figure}

In this approximation the transition from static to Coulomb+viscous friction is progressive: the friction first decreases with increasing velocity from rest state, before increasing again following the Viscous friction model. This is called the Stribeck effect, as shown in Figure \ref{fig:staticStribeckFriction}(b). In some cases, the initial friction drop can be significant, and the overall friction appears not to have the Coulomb component.

\paragraph*{Remark:} The previous models are all static, i.e. the model parameters are the same for all working regimes (joint velocities). They are unable to properly describe what happens at zero velocity (stick phase) or in the transition from stick to slip phase. Considering the joints of a humanoid robot, they operate near zero velocity or cross zero velocity quite often. In this case a dynamic or switching model is preferred. Some models, more complex, deal with hysteresis and have a local memory of previous slip/stick states, but this is out of the scope of this thesis, and will be addressed in a future work.

\subsection{The Brushless Motor Control and Modeling} \label{subsec:motor_modelling}

\paragraph*{Motor Control and PWM to Torque Transfer Function}
The BrushLess DC motor (BLDC) is a "synchronous" motor: the rotor is composed of a hub holding permanent magnets arranged in alternating pole pairs; the stator generates a magnetic field that can be rotated through electronic commutation or modulation of the current in the stator windings \citep{akin_trapezoidal_nodate}. The motor driver uses the feedback on rotor position for aligning the stator field always along the rotor quadrature axis, which results in maximum constant torque for a given PWM setpoint, and independent from the rotor angular position.
This can be achieved through Sinusoidal Commutation Control or Field Oriented Control if the loop is closed on the Park Transform of the measured phase currents \citep{kiran_field_2014} \citep{yousef_review_2015}. In this case, the motor torque , the amplitude of the rotating magnetic field, the common amplitude of the sinusoidal currents on all three phases of the stator and the motor input parameter PWM are all linearly dependent.

The rotation of the magnetic field generated by the rotor induces an electromotive force (Faraday's law) in the stator windings, $E_v$ , also called Back ElectroMotive Force (Back EMF), and therefore an induced current.
The Back EMF is proportional to fixed parameters like the number of winding turns per phase $N$, the rotor radius $r$, length $l$ and magnet flux density $B$, and to a variable parameter which is the rotor velocity $\dot{\theta}$ \citep{akin_trapezoidal_nodate}:

\begin{equation}
E_v = 2NlrB \dot{\theta} = k_{\bemf,v} \, \dot{\theta}
\end{equation}
We can then write:
\begin{equation}
\tau_{\pwm,m} = k_{\pwm,\tau} \, \PWM - k_{\bemf,\tau} \, \dot{\theta} + \tau_{f,m}
\label{equ:modelPwmTorque}
\end{equation}
Where $\PWM$ is the motor driver input parameter, $\dot{\theta}$ the rotor angular velocity, $k_{\pwm,\tau}$ is the PWM to torque coefficient and $k_{\bemf,\tau}$ is the velocity to Back EMF torque coefficient. Here we added a residual generalized friction term $\tau_{f,m}$ which could be the friction on the axis bearings. We can assume $\tau_{f,m}$ to be negligible compared to the Back EMF counter torque which can be considered as a viscous friction, since it acts as a counter torque proportional to the rotor velocity.

\subsection{Velocity, Torque and Power Conversions by Reduction or Coupling Drives}

\subsubsection{Harmonic Drives} \label{backgrnd-subsubsec:velTorqPwrConversionByHD}

Close after the motor in the actuation chain, the harmonic drive is the next component to transform the transmission dynamics variables, namely the angular velocity, the torque and the rotor apparent inertia in the sub-system [motor]-[harmonic drive]. We describe in \ref{fig:reductionDriveDynamics} a spinning mass $m$ with the respective rotational inertia ${}_\textsc{m}\dot{h}_m$ and angular velocity ${}^\textsc{m}\omega$ both expressed on the point $M$ of the shaft. We can apply this model to an humanoid robot main joint motor group, where $G$ is the harmonic drive and $m$ is the lumped mass of the fast rotating parts---the rotor and the harmonic drive wave generator. We observe a torque ${}_\textsc{m}\tau$ across the point $M$ of the shaft. The harmonic drive transforms the three dynamics quantities ${}_\textsc{m} \dot{h}_m$, ${}^\textsc{m}\omega$ and ${}_\textsc{m}\tau$ depending on the step-down ratio $\rho$, such that from its output point of view, i.e. on point $G$, the mass $m$ is seen rotating at an angular velocity ${}^\textsc{g}\omega = \rho^{-1} \, {}^\textsc{m}\omega$, the output torque is ${}_\textsc{g}\tau = \rho \; {}_\textsc{m}\tau$, and the rotational inertia is ${}^\textsc{g} \dot{h}_m = \rho^2 \, {}_\textsc{m} \dot{h}_m$. We give a more detailed explanation on these conversions further on.

\begin{axiom}
A reduction drive doesn't inject power into a mechanical system: it is conservative with respect to the power of the torques applied to input shaft, except for the power loss due to friction and the potential power stored in the elastic deformation of the drive components. The same principle is applicable to cable driven coupling mechanisms like the differential drives on humanoid robot.
\end{axiom}

\begin{note}
In the harmonic drive, the friction lies in the contact between the wave-generator and the flex-spline, and in the contact between the flex-spline and the circular spline gear teeth.
\end{note}

\paragraph*{Conversion of angular velocity and torque:}
We write the identity between the input power and the output power. We consider the effect of the generalized friction in the reduction drive as a negative power added to the input:
\begin{figure}[!t]
\centering
\includegraphics[width=0.7\columnwidth]{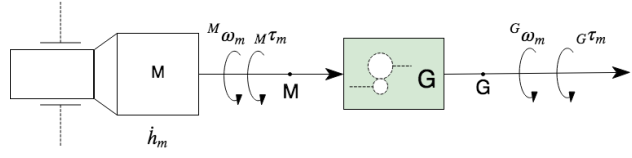}
\caption{Reduction drive dynamics conversions.} \label{fig:reductionDriveDynamics}
\end{figure}

\begin{align}
&W_{in} + W_{friction} = W_{out} \\
\Leftrightarrow \; &{}_\textsc{m}\tau \, {}^\textsc{m}\omega + {}_\textsc{m}\tau_{f,g} \, {}^\textsc{m}\omega = {}_\textsc{g}\tau \, {}^\textsc{g}\omega,
\label{equ:powerConservation}
\end{align}

\noindent{and in view of the unaltered input to output velocity ratio, we get:}
\begin{align}
&{}^\textsc{g}\omega = \rho^{-1} \; {}^\textsc{m}\omega \; \Rightarrow \;
{}_\textsc{g}\tau = \rho \; {}_\textsc{m}\tau + \rho \; {}_\textsc{m}\tau_{f,g} \label{equ:reducDrvVelConversion} \\
\Leftrightarrow \; &{}_\textsc{g}\tau = \rho \; {}_\textsc{m}\tau + {}_\textsc{g}\tau_{f,g} \label{equ:reducDrvTorqConversion} \\
\Leftrightarrow \; &{}_\textsc{g}\tau = \left( \rho + \frac{{}_\textsc{g}\tau_{f,g}}{{}_\textsc{m}\tau} \right) {}_\textsc{m}\tau
\end{align}

Where ${}_\textsc{m}\tau_{f,g}$ is the generalized friction in the reduction drive expressed on $M$ and ${}_\textsc{g}\tau_{f,g}$ the generalized friction expressed on $G$. We can either consider that the resulting input to output torque ratio is changed from $\rho$ to $\left( \rho + \frac{{}_\textsc{g}\tau_{f,g}}{{}_\textsc{m}\tau}\right) < \rho$ or consider an ideal friction-less reduction drive with torque ratio $\rho$ in series with a friction brake applying ${}_\textsc{g}\tau_{f,g}$.

\paragraph*{Conversion of angular momenta:}
Since the derivative of an angular momentum is equivalent to a torque, it is converted by the gearbox ratio the same way, as follows:
\begin{align}
&{}_\textsc{g}\dot{h}_m = \rho \; {}_\textsc{m}\dot{h}_m = \rho \; \left( {}_\textsc{m}I_m \; {}^\textsc{m} \dot{\omega} \right) \nonumber \\
\Leftrightarrow \; &{}_\textsc{g}I_m \; {}^\textsc{g} \dot{\omega} = \rho \; \left( {}_\textsc{m}I_m \; \rho \; {}^\textsc{g} \dot{\omega} \right) \nonumber \\
\Leftrightarrow \; &{}_\textsc{g}I_m = \rho^2 \; {}_\textsc{m}I_m \label{equ:reducDrvInertiaConversion}
\end{align}

Which makes ${}_\textsc{g}I_m$ be the apparent inertia of the spinning mass $m$. We then realize that in the case of a motor group on a humanoid robot, even when the standalone inertia of the rotor is negligible, it might not be the case of its apparent inertia if the actuation chain is using a high ratio gearbox and performing fast rotations: ${}_\f{g}I_m$ is then four orders of magnitude greater than ${}_\f{m}I_m$.

\paragraph*{Static friction and stiffness in Harmonic drives}
\label{backgrnd-par:staticFrictionAndStiffnessInHarmonicDrives}
In the study \citep{chedmail1970characterization}, the authors characterized the friction and stiffness in harmonic drives by running load tests on the drives, and observe that the static friction, referred to in the article as ``dry friction torque", depended on the applied load and on the angular position of the  rotating shaft. This dependency was highlighted by the hysteresis in the experimental results. They defined multiple hypothetical mechanical models of the harmonic drive in order to match the experimental observations. The models were always composed by three dynamic parameters: the friction between the gear teeth; the stiffness of the flexspline; and the play in the gears. These parameters were then combined in different, sequential arrangements: friction (F) - play (P) - Stiffness (S); F-S-P; S-P-F; and so on. They concluded verifying their initial observations and retained the model best matching the experimental results: stiffness-friction-play, with an almost null play and a friction torque of $2.6 \si{\newton\metre}$.

\Needspace{4\baselineskip}
\subsubsection{Differential Cable Drives} \label{backgrnd-subsub:diffCableDrives}

\begin{figure}[t!]
\centering
\includegraphics[width=\columnwidth]{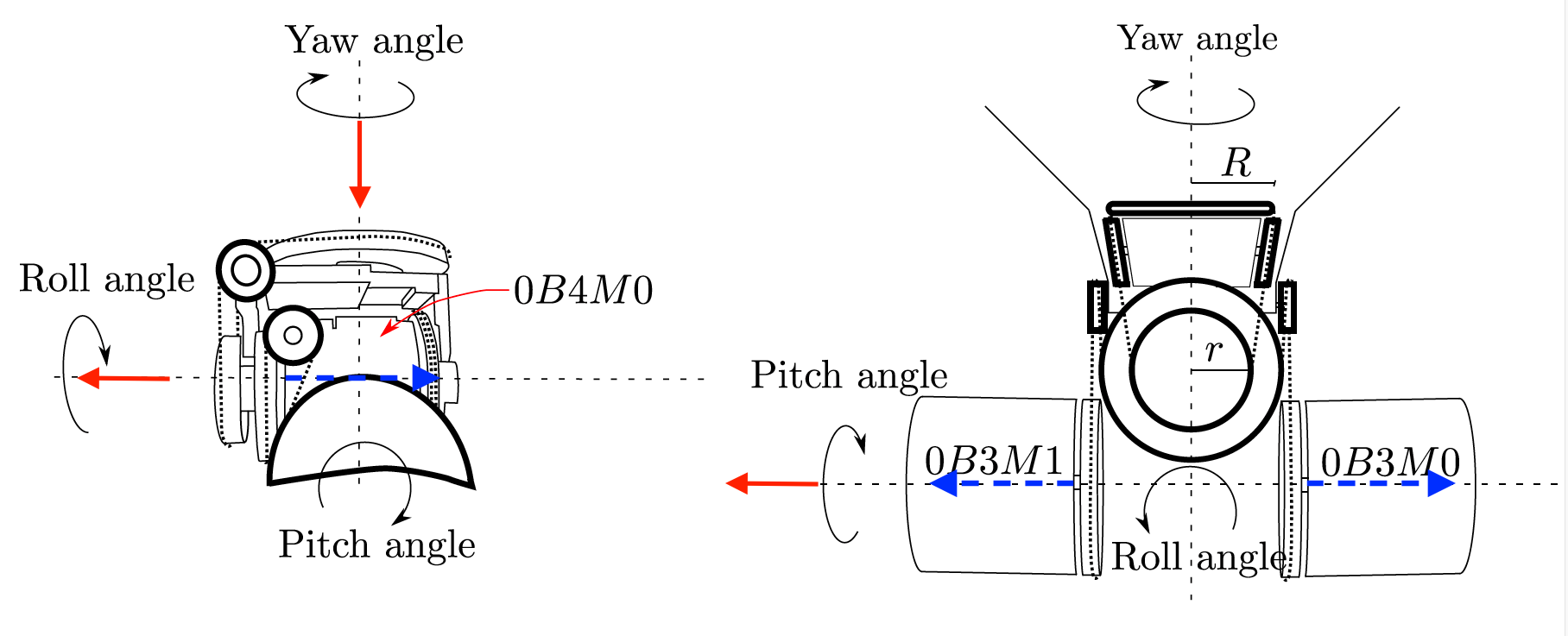}
\caption{\label{fig:sketchDifferentialDriveWaist} Differential drive coupling on the waist joints: the yaw, roll and pitch joints are actuated by the motors 0B4M0, 0B3M0,and 0B3M1 in differential configuration. Source: \citep[Fig.5, 6]{Nori2015forceReglFrontiers}.}
\end{figure}

\paragraph*{Power, velocity and torque conversion:}
Differential drives, when used, are commonly placed at the front end of a transmission chain for routing and coupling the transmission power from a set of motors to a set of joints. They can be found on a humanoid robot for coupling the waist yaw roll pitch joints to three motors 0B4M0, 0B3M0 and 0B3M1 as shown in Figure \ref{fig:sketchDifferentialDriveWaist} and illustrated in the functional sketch \ref{fig:sketchDifferentialDriveWaist}. The same principles apply here as for the reduction drives. We define the vectors of motor velocities $\mathbf{\omega_m}$ and torques $\mathbf{\tau}_m$ as being the drive input, and the vectors of joint velocities $\mathbf{\omega_j}$ and torques $\mathbf{\tau}_j$ as being the drive output. The mapping between motor and joint velocities is defined through a bijective linear transformation $T: \mathbf{\omega_m} \in \mathbb{R}^N \mapsto \mathbf{\omega_j} \in \mathbb{R}^N$, as done in \citep[3.3.3]{Nori2015forceReglFrontiers}. $T$ can be expressed as an invertible coupling matrix defined as follows:
\begin{equation}
\mathbf{\omega_j} = T \; \mathbf{\omega_m}, \label{equ:couplingVelConv}
\end{equation}

The identity between the input and output mechanical power still holds \citep[section 3.3.3]{Nori2015forceReglFrontiers}:
\begin{align}
&\mathbf{\tau}_j^\top \; \mathbf{\omega_j} = \mathbf{\tau}_m^\top \; \mathbf{\omega_m} \quad \forall \mathbf{\omega_j}, \mathbf{\omega_m} \nonumber \\
\Longrightarrow \;
&\mathbf{\tau}_j^\top \; \mathbf{\omega_j} = \mathbf{\tau}_m^\top \; T^{-1} \, \mathbf{\omega_j}  \nonumber \\
\Longleftrightarrow \; 
&\mathbf{\tau}_j^\top = \mathbf{\tau}_m^\top \; T^{-1}  \nonumber \\
\Longleftrightarrow \; 
&\boxed{\mathbf{\tau}_j = T^{- \top} \mathbf{\tau}_m}. \label{equ:couplingTorqConv}
\end{align}

\smallskip

\paragraph*{Static friction and stiffness}
We will not analyze the effect of the cables stiffness or the friction in the differential drive, instead we will assume that effect can be modeled the same way we did for the harmonic drives model, only this time with higher static friction and lower stiffness.

\section{METHODOLOGY} \label{sec:methodology}

\subsection{Assembling the Joint Actuation Model}

\begin{figure}[t]
\centering
\includegraphics[width=0.7\columnwidth]{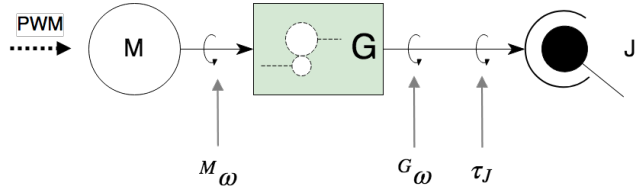}
\caption{Kinematic chain representing a single joint.} \label{fig:singleJointKinChain}
\vspace{1cm}
\end{figure}

We consider a joint actuation chain composed of a single motor \textbf{M}, a gearbox (or harmonic drive) \textbf{G} , and a rotary joint, as shown in Figure \ref{fig:singleJointKinChain}. A $\PWM$ input voltage feeds the motor, and the actuation chain delivers a joint torque $\tau_J$. For building the full joint transfer function mapping the input PWM to the output joint torque, we concatenate the sub-models described in the previous section---the motor model (\ref{equ:modelPwmTorque}) and the harmonic drive model (\ref{equ:reducDrvVelConversion}, \ref{equ:reducDrvTorqConversion}, \ref{equ:reducDrvInertiaConversion})---and integrate them in the system Euler equation.

\subsubsection{Angular momenta and Euler equation} 
\begin{figure}[h!]
\centering
\includegraphics[width=0.9\columnwidth]{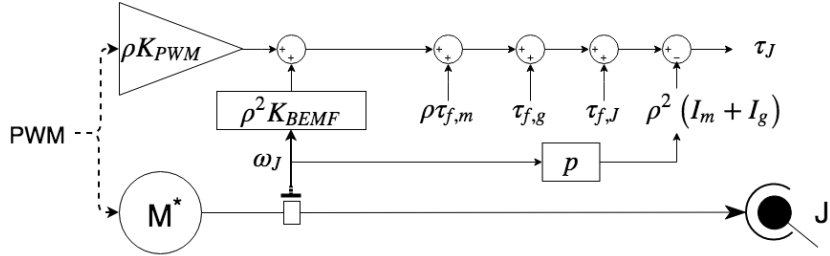}
\caption{Parallel representation of the physical model of a single joint actuation and the respective transfer function $\PWM \mapsto \tau_\textsc{j}$ in the Laplace domain. $M^*$ represents the motor block grouping the brush-less motor and the harmonic drive.} \label{fig:transferFunctionSingleJoint-2}
\end{figure}

We can write the Euler dynamics equation for the system, which equates the sum of torques applied on the system to the time derivative of the added angular momenta of the rotating elements. In this analysis we consider the child link rotating with the joint as an external element with respect to the joint actuation system:

\begin{equation}
\tau_{\pwm,m} + \tau_{f,g} + \tau_{f,\textsc{j}} - \tau_\textsc{j} = (I_m + I_g)\dot{\omega}
\end{equation}

where $I_m$ and $I_g$ are respectively the rotational inertia of the motor and the gearbox projected on the rotation axis such that $I_m \; \dot{\omega}_g$ and $I_g \; \dot{\omega}$ are scalars. In the above equation, all the derivatives are defined with respect to the same inertial frame, and expressed on the same point of the axis $G$ at the reduction drive output.
In view of \ref{equ:reducDrvVelConversion}, \ref{equ:reducDrvTorqConversion} and \ref{equ:reducDrvInertiaConversion}, we expand the motor torque and reorder the terms as illustrated in Figure \ref{fig:transferFunctionSingleJoint-2}, and rewrite the equation solving for $\tau_J$: 

\begin{align}
&\tau_{\pwm,m} + \tau_{f,g} + \tau_{f,\f{j}} - \left( I_m + I_g \right) \dot{\omega}_\f{j} = \tau_\textsc{j} \\
\begin{split}
\Longleftrightarrow &\rho k_{\pwm,\tau} \, \PWM - \rho^2 k_{\bemf,\tau} \, \omega_\textsc{j} + \rho \tau_{f,m} \\
&+ \tau_{f,g} + \tau_{f,\textsc{j}} - \rho^2 \left( I_m + I_g \right) \, \dot{\omega}_\f{j} = \tau_\textsc{j}
\end{split} \label{equ:expandedJointModelEulerEquation}
\end{align}

\subsubsection{Equivalent Motor PWM Torque}

We define the \emph{motor PWM torque} as the torque produced after the PWM input parameter and applied on the rotor, not accounting for any friction effect of mechanic or electric nature, i.e. the linear term $\rho K_{\pwm,\tau} \, \PWM = K_{\pwm,\tau}^* \, \PWM$, $K_{\pwm,\tau}^*$ being the equivalent parameter of the [motor + harmonic drive] block:

\begin{equation}
K_{\pwm,\tau}^* = \rho K_{\pwm,\tau} \label{equ:motorGroupEquivalentKpwm}
\end{equation}

\subsubsection{Equivalent Friction model}
\label{subsubsec:method_frictionModel1}

We can see that a series of terms contribute to the overall friction torque applied to the shaft: the friction on the rotor bearings $\rho \tau_{f,m}$, the Back EMF torque $\rho^2 k_{\bemf,\tau} \, \omega_\textsc{j}$, the  friction in the reduction drive $\tau_{f,g}$, and the friction on the joint bearings $\tau_{f,\f{j}}$. The approach adopted in this methodology consists in grouping all these friction terms in a single friction model, which could be the Static/Coulomb/Viscous or the Stribeck friction model.

All the terms are grouped in a single Static friction term $\tau_s$, a single Coulomb friction term $\tau_c$ and a single viscous friction term $\tau_v$:
all the friction on the bearings contributes to the terms $\tau_c$ and $\tau_v$, although we could consider them negligible with respect to the friction in the harmonic drive and the Back EMF; the Back EMF in particular is a linear function of $\omega_\textsc{j}$ and so has the properties of a viscous friction; the friction in the harmonic drive contributes to the three terms $\tau_s$, $\tau_c$ and $\tau_v$, with a significant static friction component as we will observe in the experiment results.
\begin{assumption}
For professional grade bearings, or at least in the case of iCub's brushless motors and joints bearings, the static friction on those bearings is negligible with respect with all the other friction terms and will be ignored.
\end{assumption}
We can write the three terms $\tau_c$, $\tau_v$ and $\tau_s$ as follows:
\begin{equation}
\begin{split}
\tau_c &= \tau_{c,m} + \tau_{c,g} + \tau_{c,\f{j}} = - K_c \, \sign(\omega_\f{j}) \\
\tau_v &= \tau_{v,m} + \tau_{\textsc{bemf}} + \tau_{v,g} + \tau_{v,\f{j}} = - K_v \, \omega_\f{j} \\
\tau_s &= \tau_{s,g} (\omega_\f{j},\sigma_s) \\
\tau_f (\omega_\f{j},\sigma_s) &= \tau_c (\omega_\f{j}) + \tau_v (\omega_\f{j}) + \tau_s (\omega_\f{j},\sigma_s),
\end{split} \label{equ:frictionEquivalentParams}
\end{equation}

Where $\tau_{c,\cdot}$, $\tau_{v,\cdot}$ and $\tau_{s,\cdot}$ relate to the nature of the friction---respectively Coulomb, viscous and static---and $\tau_{\cdot,m}$, $\tau_{\cdot,g}$, $\tau_{\cdot,\f{j}}$ relate to the origin of the friction---respectively the motor bearings, harmonic drive (gearbox) and joint bearings. $\sigma_s$ is the maximum static friction measured at zero velocity.

\subsubsection{Identification of The Equivalent Model Parameters}

In view of \eqref{equ:expandedJointModelEulerEquation}, \eqref{equ:motorGroupEquivalentKpwm} and \eqref{equ:frictionEquivalentParams}, we can rewrite the output joint torque as:

\begin{equation}
{}_\f{g}\tau_\f{j} = K_{\pwm}^* \; \PWM - \left({}_\f{g}I_m + {}_\f{g}I_g \right)  {}^\f{g}\dot{\omega} - K_c \, \sign({}^\f{g}\omega) - K_v \, {}^\f{g}\omega + \tau_s
\label{equ:jointTorqueDecomposition1}
\end{equation}

The identification of the motor and gearbox inertial parameters $I_m + I_g$ is out of the scope of this thesis, so we assume they are given. We then need to identify the motor parameter $K_{\pwm}^*$ and the friction parameters $\tau_s$, $K_c$ and $K_v$.

The identification will be performed in two phases: the first phase identifies the viscous friction parameters $\tau_s$, $K_c$ and $K_v$; the second phase identifies the motor parameter $K_{\pwm}^*$. This approach allows to simplify the fitting of the model parameters by reducing the problem dimension for each phase of the identification.

\subsection{First phase - friction parameters identification}
\label{method-subsec:methodFirstPhase}

\subsubsection{Symmetric Coulomb/viscous Friction Model} \label{method-subsubsec:symmCoulobViscousModel1}

We initially only consider the Coulomb and viscous friction components, for describing the base estimation algorithm, and will later introduce the static friction component. We need to place the system in a condition where only the friction torques are present and acting on the shaft. This is achieved by setting the motor input $\PWM$ to zero, canceling the motor PWM torque and leaving only the motor internal mechanical friction and the counter torque due to the Back EMF. In normal operating conditions, the Back EMF adds a significant contribution to the overall joint actuation friction, and for this reason, it's crucial to account for that contribution in this phase of the identification. The Back EMF creates induced currents in the motor electromagnetic coils circuit \footnote{located in the stator in the case of brushless motors} which result in a torque opposed to the motor PWM torque.  Setting the $\PWM$ parameter as mentioned above, instead of turning off the motor, allows to keep the stator circuit closed and the Back EMF current to flow. In view of \eqref{equ:jointTorqueDecomposition1} and $\PWM = 0$, we express the joint torque as follows:

\begin{equation}
{}_\f{g}\tau_\f{j} + \left({}_\f{g}I_m + {}_\f{g}I_g \right)  {}^\f{g}\dot{\omega} = - K_c \, \sign({}^\f{g}\omega) - K_v \, {}^\f{g}\omega \label{equ:kc-kv-model}
\end{equation}

Where the joint velocity ${}^\f{g}\omega$ and acceleration ${}^\f{g}\dot{\omega}$ are respectively measured from joint encoders or inertial sensors. The joint torque ${}_\f{g}\tau_J $ is measured by a joint torque sensor or estimated from Force-Torque sensors measurements and a modified inverse dynamics algorithm as seen in \citep[Chapter 4 section 4.4.2]{traversaro2017modelling}. We get the training data from a set of measurements, and then fit the model \eqref{equ:kc-kv-model} by defining and solving the over-constrained linear system below:

\begin{equation}
X \, \Theta = \mathbf{y} \label{equ:ident-linear-system}
\end{equation}
With,
\begin{align}
\begin{split}
&\mathbf{y} = \mathbf{{}_\f{g}\tau_\f{j}} + \left({}_\f{g}I_m + {}_\f{g}I_g \right)  \mathbf{{}^\f{g}\dot{\Omega}} \\
&X = \begin{bmatrix}
\text{sign}(\omega_{1})   & \text{sign}(\omega_{2})   & \dots  & \text{sign}(\omega_{n}) \\
\omega_{1} & \omega_{2} & \dots  & \omega_{n}
\end{bmatrix}^{\top}\\
&\Theta = \begin{bmatrix}
- K_c & - K_v
\end{bmatrix}^{\top}
\end{split}
\label{equ:linearSysFrictionEst1}
\end{align}

where $\omega_{i}$ is the measured velocity at instant $i$, $\mathbf{{}^\f{g}\dot{\Omega}}$ is the column vector of the joint angular acceleration measurement samples and $\mathbf{{}_\f{g}\tau_\f{j}}$ is the column vector of the joint torque measurement samples. Obtaining the system \eqref{equ:linearSysFrictionEst1} is straightforward as it's just the matrix formulation of \eqref{equ:kc-kv-model}. The linear least squares solution can be computed by inverting the matrix $X$ through a Moore-Penrose left pseudo-inverse:
\begin{equation}
\Theta = \left( X^{\top} X \right)^{-1} X^{\top} \, \mathbf{y}
\end{equation}
$X$ is of full column rank since the second column is an evenly distributed set of velocities and so cannot be a multiple of the first column which is a series of $\pm 1$ elements. The matrix $X^\top X$ is well conditioned, assuming that we use a reasonable range of joint velocities and have a low level of noise in the encoder measurements. In any case, we verify numerically the condition number of that matrix in the experiments in section \ref{sec:experiments}.

\subsubsection{Stribeck/Coulomb/viscous friction model}
\label{method-subsubsec:stribeckCoulombViscousModel3}

If the static friction in the reduction drive is significant, for instance around $2 \, \si{\newton\metre}$ or higher, the results obtained with the models proposed above will not result in an accurate fitting, giving a good approximation at higher velocities, and a poor approximation near zero velocities, where the friction torque curve appears to be quite non linear, and right before the state transitions to a ``stick" phase, as we will observe in the experiment results (\ref{subsec:exp-frictionParamsEst}). The non linearity and the unaccounted effects of the ``slip-stick" phase transitions are aggravated when the joint reaches or crosses the zero velocity.

For that reason we now integrate in the estimation algorithm the model combining the Coulomb and Viscous friction components with the Stribeck effect $\tau_s(\omega)$ \cite{van2009study} \cite{4653107}. The Stribeck effect is modeled here with an empiric non linear function, but still linear with respect to the fitting parameters. The function we are looking for has to approximate the torque response characterized by the plot in Figure \ref{fig:r-knee-friction-slow-model-kckv}, which strongly resembles the Stribeck model depicted in \ref{fig:staticStribeckFriction} and analyzed closer in this section. For that and in view of \eqref{equ:frictionEquivalentParams}, the overall friction function $\tau_f(\omega)$ has to meet the following properties:
\begin{align*}
&\lim_{\omega \rightarrow 0^-}{\tau_f(\omega)} = \sigma_s^- &\lim_{\omega \rightarrow +\infty}{\tau_f(\omega)} = \tau_c(\omega) + \tau_v(\omega) \\
&\lim_{\omega \rightarrow 0^+}{\tau_f(\omega)} = \sigma_s^+ &\lim_{\omega \rightarrow -\infty}{\tau_f(\omega)} = \tau_c(\omega) + \tau_v(\omega)
\end{align*}
By applying \eqref{equ:frictionEquivalentParams} in each of the above equations we get:
\begin{align}
&\lim_{\omega \rightarrow 0^-}{\tau_s(\omega)} = \sigma_s^- - \tau_c(\omega) - \tau_v(\omega) = \sigma_s^- - \tau_c(\omega) \\
&\lim_{\omega \rightarrow 0^+}{\tau_s(\omega)} = \sigma_s^+ - \tau_c(\omega) \\
&\lim_{\omega \rightarrow \infty}{\tau_s(\omega)} = 0
\end{align}
We could have chosen for $\tau_s(\omega)$ an hyperbolic function $\propto \frac{1}{\omega^n}$ but such function would have a singularity at $\omega = 0$ that could be avoided with the shifted function: $\tau_s(\omega) \propto \frac{1}{(\omega-\omega_0)^n}$, only this introduces a new parameter $\omega_0$ with respect to which $\tau_s$ would be non-linear. The good candidate we found is an exponential function linearly parameterized. Let's consider the function $g(x,\sigma) = \left( \sigma - K_c \right) \, e^{-x}$. $\tau_s(\omega)$ is symmetric with respect to the origin, coincides with $g(\omega,\sigma_s^+)$ in the half-plane $\omega>0$, and coincides with $-g(-\omega,\sigma_s^-)$ in the half-plane $\omega<0$, as illustrated in figure \ref{fig:stribeckDecomposition}. We can then write:
\begin{figure}[t!]
\centering
\includegraphics[width=0.9\columnwidth]{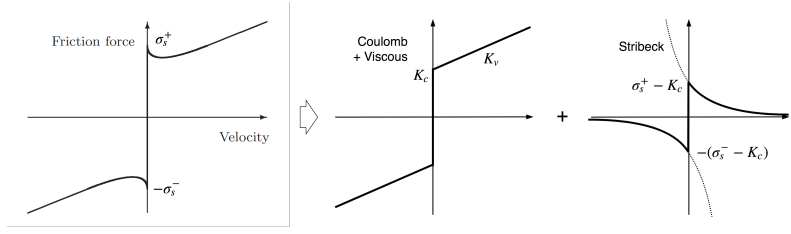}
\caption{Decomposition of Stribeck model into Coulomb/Viscous + exponential function.} \label{fig:stribeckDecomposition}
\end{figure}
\begin{align}
\begin{split}
\tau_s(\omega) &= u^+(\omega) \left( \sigma_s^+ - K_c \right) \, e^{-\omega}
- u^+(-\omega) \left( \sigma_s^- - K_c \right) \, e^{\omega} \\
&= - \left( u^+(\omega) \, e^{-\omega} + u^-(\omega) \, e^{\omega} \right) K_c \\
& \quad + u^+(\omega) \, e^{-\omega} \, \sigma_s^+ \\
& \quad + u^-(\omega) \, e^{\omega} \, \sigma_s^-
\end{split}
\end{align}
where $u^+$ , $u^-$, $r^+$ and $r^-$ are respectively the step and ramp functions such that:
\begin{align}
u^+(x)&:=\begin{cases}
0, & x < 0 \\
1, & x  \geq 0
\end{cases} \quad
u^-(x):=\begin{cases}
-1, & x < 0 \\
0, & x  \geq 0
\end{cases} \\
r^+(x)&:=\begin{cases}
0, & x < 0 \\
x, & x  \geq 0
\end{cases} \quad
r^-(x):=\begin{cases}
-x, & x < 0 \\
0, & x  \geq 0
\end{cases}
\end{align}
So the overall friction can be posed as the linear system:
\begin{align}
\begin{split}
&\mathbf{y} = \mathbf{{}_\f{g}\tau_J} + \left({}_\f{g}I_m + {}_\f{g}I_g \right)  \mathbf{\dot{\Omega}_\f{g}} \\
&X = X_1 + X_2 \\
&\Theta = \begin{bmatrix}
-K_c & -K_v & -\sigma_s^+ & -\sigma_s^-
\end{bmatrix}^{\top},
\end{split}
\label{equ:linearSysFrictionEst2.1}
\end{align}
\begin{flushleft}
with $X_1$ and $X_2$ defined as follows:
\end{flushleft}
\begin{align}
\begin{split}
X_1 &= \begin{bmatrix}
u^+(\omega_{1}) \left( 1+e^{-\omega_{1}} \right) & \hdots & u^+(\omega_{n}) \left( 1+e^{-\omega_{n}} \right) \\
r^+(\omega_{1}) & \hdots & r^+(\omega_{n}) \\
u^+(\omega_{1}) \, e^{-\omega_{1}} & \hdots & u^+(\omega_{n}) \, e^{-\omega_{n}} \\
0 & \hdots & 0
\end{bmatrix}^{\top} \\
&= \diag \left( u^+(\omega_{1}) \hdots u^+(\omega_{n}) \right)
\begin{bmatrix}
\left( 1+e^{-\omega_{1}} \right) & \hdots & \left( 1+e^{-\omega_{n}} \right) \\
\omega_{1} & \hdots & \omega_{n} \\
e^{-\omega_{1}} & \hdots & e^{-\omega_{n}} \\
0 & \hdots & 0
\end{bmatrix}^{\top}
\end{split} \label{equ:linearSysFrictionEst2.2} \\
\begin{split}
X_2 &= \begin{bmatrix}
u^-(\omega_{1}) \left( 1+e^{\omega_{1}} \right) & \hdots & u^-(\omega_{n}) \left( 1+e^{\omega_{n}} \right) \\
r^-(\omega_{1}) & \hdots & r^-(\omega_{n}) \\
0 & \hdots & 0 \\
u^-(\omega_{1}) \, e^{\omega_{1}} & \hdots & u^-(\omega_{n}) \, e^{\omega_{n}}
\end{bmatrix}^{\top} \\
&= \diag \left( u^-(\omega_{1}) \hdots u^-(\omega_{n}) \right)
\begin{bmatrix}
\left( 1+e^{\omega_{1}} \right) & \hdots & \left( 1+e^{\omega_{n}} \right) \\
\omega_{1} & \hdots & \omega_{n} \\
e^{\omega_{1}} & \hdots & e^{\omega_{n}} \\
0 & \hdots & 0
\end{bmatrix}^{\top} \label{equ:linearSysFrictionEst2.3}
\end{split}
\end{align}

\subsection{Second phase - Motor Parameter Identification}
\label{method-subsec:methodSecondPhase}

\paragraph*{\textbf{Algorithm}}

The friction coefficients having been identified, we can now use them in the joint dynamics equation \eqref{equ:jointTorqueDecomposition1} for identifying $K_{\pwm}^*$. The approach is then equivalent, defining a system correlating this time the input duty cycle $\PWM$ and the output joint torque $\tau_\f{j}$, where $K_{\pwm}^*$ is the only unknown parameter. From \eqref{equ:jointTorqueDecomposition1} we can write \eqref{equ:ident-linear-system} with:
\begin{align}
\begin{split}
&\mathbf{y} = \mathbf{{}_\f{g}\tau_\f{j}} + \left( {}_\f{g}I_m + {}_\f{g}I_g \right)  \mathbf{{}^\f{g}\dot{\Omega}} + K_c \, \mathbf{\sign({}^\f{g}\Omega)} + K_v \, \mathbf{{}^\f{g}\Omega} - \mathbf{{}_\f{g}\tau_s} \\
&X = \begin{bmatrix}
1 & 1   & \dots  & 1 \\
\textsc{PWM}_{1} & \textsc{PWM}_{2} & \dots  & \textsc{PWM}_{n}
\end{bmatrix}^{\top}\\
&\Theta = \begin{bmatrix}
\tau_0 & K_{\pwm}^*
\end{bmatrix}^{\top},
\end{split}
\label{equ:linearSysPwmEst1}
\end{align}

where $\mathbf{\sign({}^\f{g}\Omega)} = \begin{bmatrix} \sign(\omega_{1}) \hdots \sign(\omega_{n}) \end{bmatrix}^\top$. $\tau_0$ accounts for an eventual offset current in the stator windings. A null motor current and torque could then be obtained from an input value $\PWM = - K_{\pwm}^{-1} \, \tau_0$. The system is solved as the previous one, through the computation of a Moore-Penrose left pseudo-inverse.

\subsection{The Coupled Joints Case}
\label{subsec:method_jointCoupling}

The kinematic and dynamic coupling implemented by the differential drives has to be accounted for when correlating joint with motor torques. this aspect significantly impacts the measurement procedures and the estimation results. Let us consider three joints $J_1, J_2, J_3$ actuated by three motors $M_a, M_b, M_c$ through a differential coupling. We depict in Fig. \ref{fig:coupledJointKinChain} the respective actuation system.

\subsubsection{Velocities and torques transformation}

\begin{figure}[t]
\centering
\includegraphics[width=0.5\columnwidth]{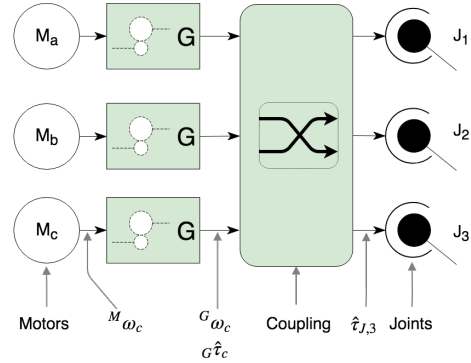}
\caption{Kinematic chain representing three coupled joints, the gearboxes and motors. ${}^\f{m} \omega_c$, ${}^\f{g} \omega_c$ and ${}_\f{g} \tau_c$ are respectively shaft angular velocities and the measured or estimated torque on the motor C transmission chain. $\tau_{\f{j},3}$ is the output joint torque estimated on the joint $J_3$.} \label{fig:coupledJointKinChain}
\end{figure}

We've seen in section \ref{backgrnd-subsub:diffCableDrives} how to transform velocities \eqref{equ:couplingVelConv} and torques \eqref{equ:couplingTorqConv} between the input and the output of a coupling system like the differential drive on the torso. If we apply \eqref{equ:couplingVelConv} and \eqref{equ:couplingTorqConv} to the quantities illustrated in the figure \ref{fig:coupledJointKinChain}, we get:

\begin{align}
\omega_\f{j} &= T \, {}^\f{g}\omega, \quad \text{with} \quad \omega_\f{j} = 
\begin{bmatrix}
\omega_{\f{j},1} \\ \omega_{\f{j},2} \\ \omega_{\f{j},3}
\end{bmatrix}, \quad {}^\f{g}\omega =
\begin{bmatrix}
{}^\f{g}\omega_a \\ {}^\f{g}\omega_b \\ {}^\f{g}\omega_c
\end{bmatrix}, \\
\tau_\f{j} &= T^{-\top} \, {}_\f{g}\tau, \quad \text{with} \quad \tau_\f{j} = 
\begin{bmatrix}
\tau_{\f{j},1} \\ \tau_{\f{j},2} \\ \tau_{\f{j},3}
\end{bmatrix}, \quad {}_\f{g}\tau =
\begin{bmatrix}
{}_\f{g}\tau_a \\ {}_\f{g}\tau_b \\ {}_\f{g}\tau_c
\end{bmatrix},
\end{align}

\subsubsection{Decorrelating the coupled motors} \label{method-subsubsec:decorrelatingCoupledMotors}

At this point we have the choice between two methods: either we express all the quantities at the joint level in the algorithms proposed in this section like \eqref{equ:linearSysFrictionEst1} and \eqref{equ:linearSysPwmEst1}, either we express them at the point $G$ of each motor group shaft $(a)$, $(b)$ and $(c)$ and apply the algorithms on the torques ${}_\f{g}\tau_a$ or ${}_\f{g}\tau_b$ or ${}_\f{g}\tau_c$. This second approach is indeed simpler: we first transform all the joint velocities and torques measurements $\mathbf{\omega_\f{j}}$ and $\mathbf{\tau_\f{j}}$,  into the respective transformed quantities $\mathbf{{}^\f{g}\omega}$ and $\mathbf{{}_\f{g}\tau}$ expressed in the point $G$ of the motor group shaft $(a)$, $(b)$ and $(c)$, as illustrated in \ref{fig:coupledJointKinChain}. We then just have to apply the proposed methodology described in sections \ref{method-subsec:methodFirstPhase} and \ref{method-subsec:methodSecondPhase}, using ${}_\f{g}\tau$ in place of ${}_\f{g}\tau_\f{j}$. This converts the problem into three independent problems, each on a single motor actuation chain.

The first method is motivated by the choice to perform the estimation joint wise, each joint "seeing" a virtual motor resulting from its interaction with the coupled motors. In this case, each joint is moved alone, which results in the coupled motors moving simultaneously. The second method instead, is motivated by an estimation motor wise, which then requires to constrain the motors to move one by one and to focus on the estimation of that motor's actuation chain parameters.

The idea is to constrain all the power to be exchanged only between one motor and the three coupled joints. Let's consider the actuation chain between the motor $M_c$ and the three joints for the friction and motor parameters identification. We first block the remaining motors $M_a$ and $M_b$ in any desired position (e.g. position initially giving the origin position of the three joints). We then apply the torque transformation to get the joint torque projected on the motor $M_c$ actuation chain ${}_\f{g}\tau_c$ illustrated in Fig. \ref{fig:coupledJointKinChain} and defined as follows:

\begin{equation*}
{}_\f{g}\tau_c = T_c^\top \; \begin{bmatrix} \tau_{\f{j},1} & \tau_{\f{j},2} & \tau_{\f{j},3} \end{bmatrix} ^\top,
\end{equation*} 

where $T_c$ is the column of $T$ related to motor $M_c$. We then apply the dynamics equation \ref{equ:jointTorqueDecomposition1} to $M_c$, replacing ${}_\f{g}\tau_J$ by $\tau_{g,c}$ and $\omega_\f{g}$ by $\omega_{g,c}$. At last we solve the system for the friction and motor parameters following the algorithms proposed in \ref{method-subsec:methodFirstPhase} and in \ref{method-subsec:methodSecondPhase}. 

We followed the second method, which presents the following benefits: it is less sensitive to model uncertainties that can arise from bad estimates of the coupling reduction ratios and the cables elongation in case of a cable driven differential; it allows to track unexpected slip/stick phase transitions separately for each of the three motor actuation chains; whatever the complexity of the coupling it will be easier to cover a desired range of velocities on each motor.

\begin{note}
In platforms having Field Oriented Control motor drives, the motor velocity can be estimated from the measurements of a high resolution encoder placed on the motor stator, which benefits from much higher accuracy compared to the typical three hall-effect sensors used in motors driven by a Trapezoidal Control.
\end{note}

\section{EXPERIMENTAL RESULTS} \label{sec:experiments}

We have tested the estimation algorithms proposed in section \ref{sec:methodology}, on the robot iCub. We have estimated the joint friction and motor parameters on the right leg knee and on the coupled joints---yaw, roll, pitch---connecting the waist to the torso. The sensors data were captured at 100Hz sampling rate: joint positions, velocities, torque and motor velocities.

\subsection{First Phase - Friction Parameters Identification}
\label{subsec:exp-frictionParamsEst}

\subsubsection{Procedure}

After setting the $\PWM$ to zero, we apply an external torque to the joint in an oscillating motion \footnote{This can be done manually or with an external actuation.}. It is preferred to avoid significant accelerations in order to minimize and eventually neglect the inertial terms, otherwise we use the inertial sensors measurements for estimating and accounting for the inertial terms. We then apply the method described in \ref{method-subsubsec:symmCoulobViscousModel1} on the acquired data. When applying the model, we verified numerically that the matrix $X^\top X$ is well conditioned, which is a requirement for the pseudo-inverse computation to give an accurate solution.

\subsubsection{Experiment Results}
\label{exp-subsubsec:experimentalResults}

\begin{figure}[!t]
\centering
\includegraphics[width=0.9\columnwidth]{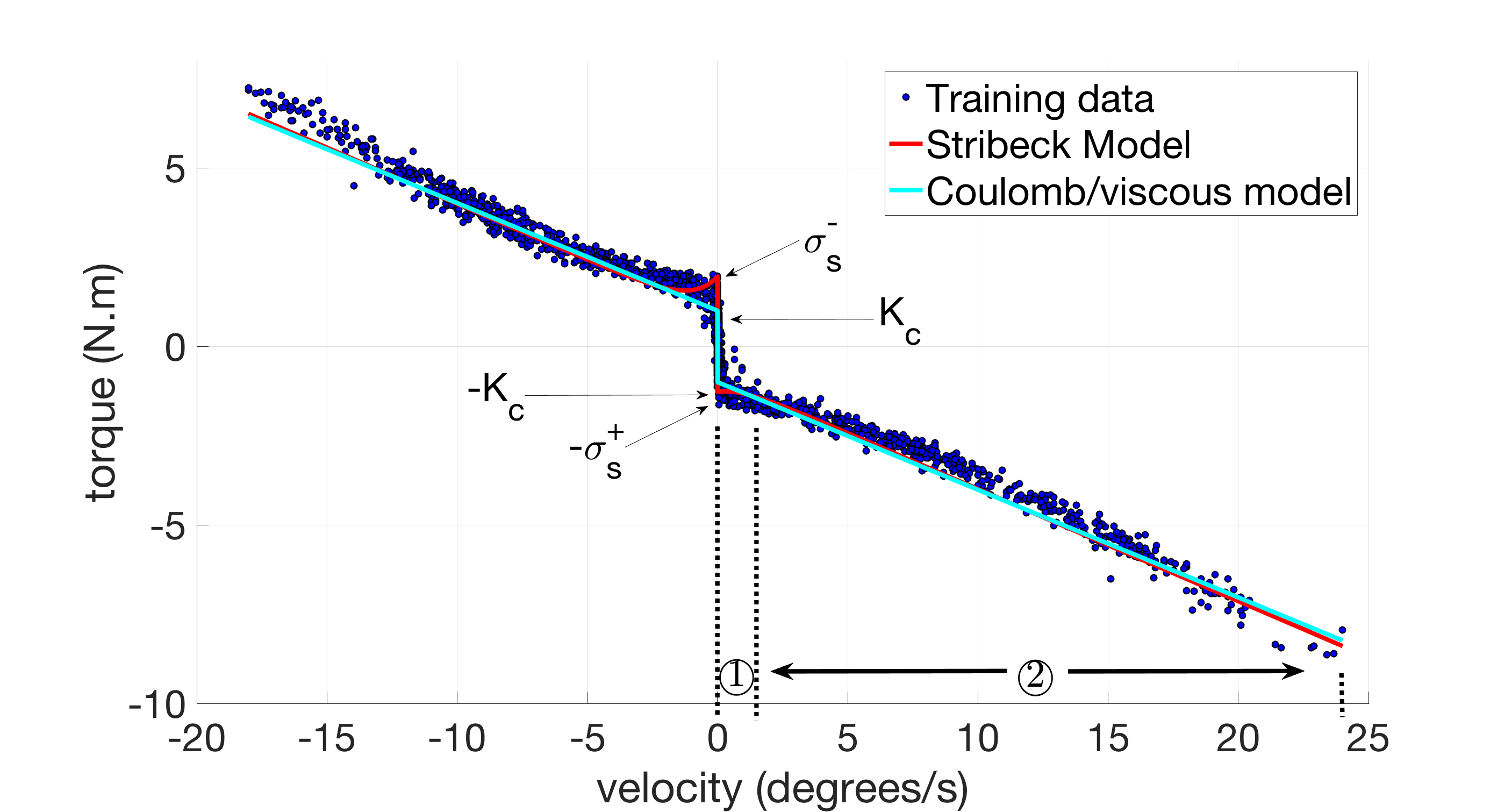}
\caption{Friction model fitting results on the right leg knee. Coulomb/Viscous model parameters: $K_c \sim 1 \si{\newton\metre}$, $K_v \sim 0.30 \si{\newton\metre\second\per\deg}$. Stribeck model parameters: static frictions $\sigma_s^+ \sim 1.27 \si{\newton\metre}$, $\sigma_s^- \sim 1.95 \si{\newton\metre}$; $K_c \sim 0.85 \si{\newton\metre}$; $K_v \sim 0.31 \si{\newton\metre\second\per\deg}$. $\textcircled{1}$: Boundary Lubrication regime, $\textcircled{2}$: Full Fluid Lubrication regime (viscous friction).} \label{fig:r-knee-friction-slow-model-kckv}
\end{figure}

The plot in Figure \ref{fig:r-knee-friction-slow-model-kckv} shows the friction estimation results on the right leg knee, a single joint free of any coupling or cable driven actuation.

As we can observe in the plot \ref{fig:r-knee-friction-slow-model-kckv}, the model (cyan color line) doesn't seem to fit well the data near the zero velocity: while at zero velocity the model gives a Coulomb friction $K_c \sim 1 \si{\newton\metre}$, the measured torque is twice as high, rising up to $\sim 2 \si{\newton\metre}$.

Later tests on most of the joints on the right leg of iCub---hip pitch/roll/yaw, knee---revealed the same consistent increase of friction when the joint velocity decreases to zero, similar to the Stribeck effect, as can be observed in all the plots of friction torque in this section. When the surfaces are in solid-to-solid contact, we are in a regime called ``Boundary Lubrication", the sheared solid surfaces being the boundary lubricants.

Unlike the Coulomb/Viscous model, the Stribeck model (as seen in the same figure) gives a good estimate of $\sigma_s^+$ and $\sigma_s^-$. Although we get approximately the same coefficient $K_v$, $K_c$ differs significantly between the two models.

\paragraph*{A better fit of the ``Boundary Lubrication":}
The constraint on the derivative of the model function at zero velocity, $\left[ \delta{\tau}_f / \delta{\dot{q}} \right]_{\dot{q}=0} = 0$, ensures that we better fit the ``Boundary Lubrication" plateau. we can observe the fitting improvement in Figure \ref{fig:r-knee-friction-slow-model-kckv}, where we applied the constraint on the right derivative, i.e. $\left[ \delta{\tau}_f / \delta{\dot{q}} \right]_{\dot{q}=0^+}$.

\bigskip

We have run a friction estimation trial on all the right leg joints, listed in the table \ref{tabl:friction-params-results-right_leg}. These results haven't been validated yet with the low level controller. On top of that, the present low level controller doesn't integrate yet static nor Coulomb friction compensation.

\subsubsection{The Coupled Joints Case}
\label{exp-subsubsec:theCoupledJointsCase-friction}

\begin{figure}[!t]
\centering
\includegraphics[width=0.9\columnwidth]{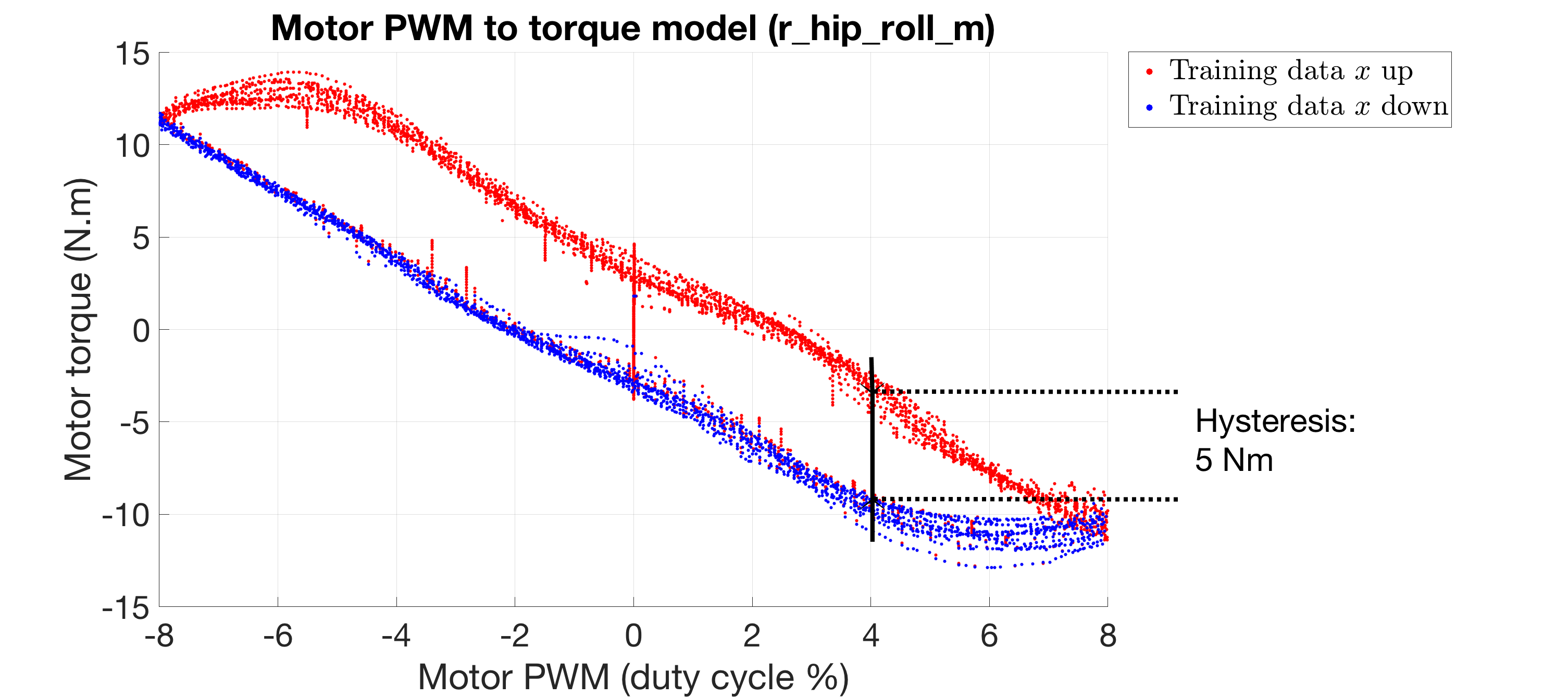}
\caption{Hysteresis in the PWM to torque characteristics due to static friction in the harmonic drive.} \label{fig:hip-roll-pwm-directctrl-hystColor-amp8-freq0p3}
\end{figure}

The PWM-to-torque identification results are illustrated in Figure \ref{fig:hip-roll-pwm-directctrl-hystColor-amp8-freq0p3}, showing the impact of the Harmonic Drive hysteresis on the measured output torque. The motor velocity-to-friction torque model identification results are equivalent to those obtained for a single joint and will not be illustrated in this paper.

\begin{table}[t]
\rowcolors{2}{gray!20}{gray!5}
\begin{tabularx}{\textwidth}{||l|X|X|X||}
\hline
Joint & $\sigma_s^+ \, (\si{\newton\metre})$ & $\sigma_s^- \, (\si{\newton\metre})$ & $K_v \, (\si{\newton\metre\second\per\deg})$ \\
\hline\hline
right hip pitch &  1.31 &  2.24 &  1.16 \\
right hip roll & 2.49 & 0.96 & 0.50 \\
right hip yaw & 1.30 &  2.58 &  1.22 \\
right knee & 1.27 &  1.95 & 0.69 \\
right ankle pitch & 2.90 & 0.71 &   1.32 \\
right ankle roll & 2.05 &  1.29 &  1.20 \\
\hline
\end{tabularx} \vspace{0.2cm}
\caption{Friction parameters extimation results on the iCub right leg joints.} \label{tabl:friction-params-results-right_leg}
\end{table}

\section{CONCLUSIONS}

The paper first defines a detailed model of a typical joint low level actuation sub-system used on a humanoid robot with electrical actuators, then extends the model to coupled joints. It then highlights how a linear PWM-to-torque model (before the conversion in the Harmonic Drive), free of ripples, depends on the motor hardware and a Field Oriented Control mode for tracking the motor current. The Back EMF and the friction in the Harmonic Drive are identified as the main sources of friction. The method uses a least squares fitting algorithm for identifying the parameters of the Stribeck model, with three dynamic regimes: elastic deformation, Boundary Lubrication, Full Fluid Lubrication. This model fits the better the measured friction torques. Unlike previous works on friction identification, the method presented here was performed in open loop control, relying on the joint torque estimations based on force-torque sensors instead of motor currents adjusted by a closed loop control. This guarantees a simpler and more reliable control of the motor input, avoiding any quantization in the motor torque response, and better revealing in the plots the static friction from the Harmonic Drive. In a future work, the identified friction parameters shall be used on a dynamic friction model.

%
%
\bibliographystyle{spmpsci}
\bibliography{root}
\end{document}